\documentclass{article}

    \PassOptionsToPackage{numbers, compress}{natbib}
    \usepackage[preprint]{neurips_2020}

\usepackage[utf8]{inputenc} 
\usepackage[T1]{fontenc}    
\usepackage{hyperref}       
\usepackage{url}            
\usepackage{booktabs}       
\usepackage{amsfonts}       
\usepackage{nicefrac}       
\usepackage{microtype}      

\usepackage{graphicx}  
\graphicspath{ {./images/} }

\usepackage{subcaption}

\usepackage[title]{appendix}

\newcommand{\comment}[1]{}

\title{A Generalizable Approach to Learning Optimizers}

\author{%
  Diogo Almeida
  \qquad
  Clemens Winter
  \qquad
  Jie Tang
  \qquad
  Wojciech Zaremba
  \\ OpenAI
}

\begin{document}

\maketitle

\begin{abstract}
A core issue with learning to optimize neural networks has been the lack of generalization to real world problems. To address this, we describe a system designed from a generalization-first perspective, learning to update optimizer hyperparameters instead of model parameters directly using novel features, actions, and a reward function. This system outperforms Adam at all neural network tasks including on modalities not seen during training. We achieve 2x speedups on ImageNet, and a 2.5x speedup on a language modeling task using over 5 orders of magnitude more compute than the training tasks.
\end{abstract}

\section{Introduction}

\begin{figure}[h]
    \centering
    \includegraphics[scale=0.33]{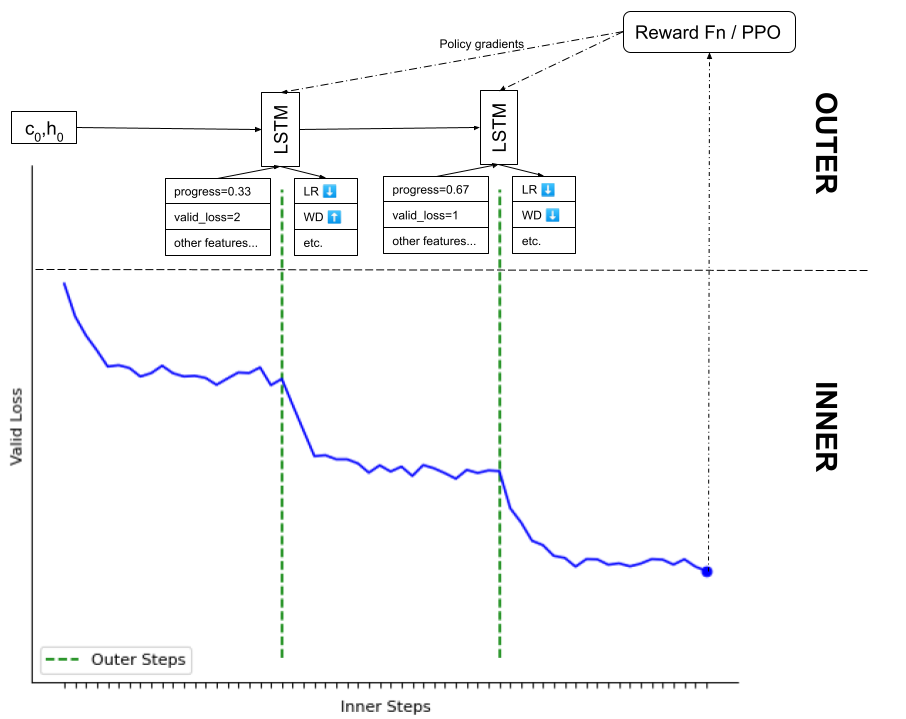}
    \caption{Simplified diagram of our LHOPT approach. Some small number of times through training, an LSTM controller is used to update optimizer hyperparameters based on input statistics from the training process. We optimize for generalization by performing hyperparameter updates instead of parameter updates, decouple update frequency from inner task size, use unitless features and relative actions, solely optimize for final performance, and use a novel reward function.}
    \label{fig:diagram}
\end{figure}

Why haven't we automated deep neural network optimization yet? We know that a problem exists: optimization settings greatly affect the quality of trained models~\citep{bengio2012practical}. We know that tools exist that can tackle problems like this: modern reinforcement learning methods have made great strides at solving difficult sequential decision making problems with imperfect information, such as Dota 2~\citep{dota2}. And we know that solving this problem would be impactful: the field is continuing to scale up models~\citep{scaling_laws} and even small relative speedups could result in saving thousands of petaflop/s-days~\citep{gpt3}.

Previous work~\citep{l2l,scale_generalize,pathologies,luke2020} has proposed solutions to the problem of learning how to optimize neural networks, but despite demonstrating large speedups on problems from or close to their training distribution, all have shown limited generalization far outside of that training distribution, and not a single one is used for real use cases. While there are many problems learned optimizers have to solve, we, like previous work~\citep{luke2020}, argue that the core issue is generalization.

In this work, we provide a description of a system as well as motivation for novel design choices for learned optimizers with the focus on generalizing to and increasing computational efficiency of real world problems. We caveat that this work is preliminary with limitations outlined throughout the paper.

\subsection{Contributions}

Most existing methods attempt to learn a general update rule from scratch: learning parameter updates~\citep{l2l,scale_generalize,pathologies,luke2020}. In contrast, we leverage the priors of existing optimization algorithms as much as possible, and instead learn hyperparameter updates. We call this class of learned optimizers Learned Hyperparameter Optimizers (LHOPTs). The resulting optimizers can be interpreted as having data-driven schedules that interpolate between hand-designed optimizers.

In addition to presenting the general framework of learning to update hyperparameters, we also revisit many design decisions for learned optimizers from a generalization-first perspective and present novel approaches for actions, features, and reward functions for learned optimizers in general.

\begin{figure}[h]
    \centering
    \includegraphics[scale=0.35]{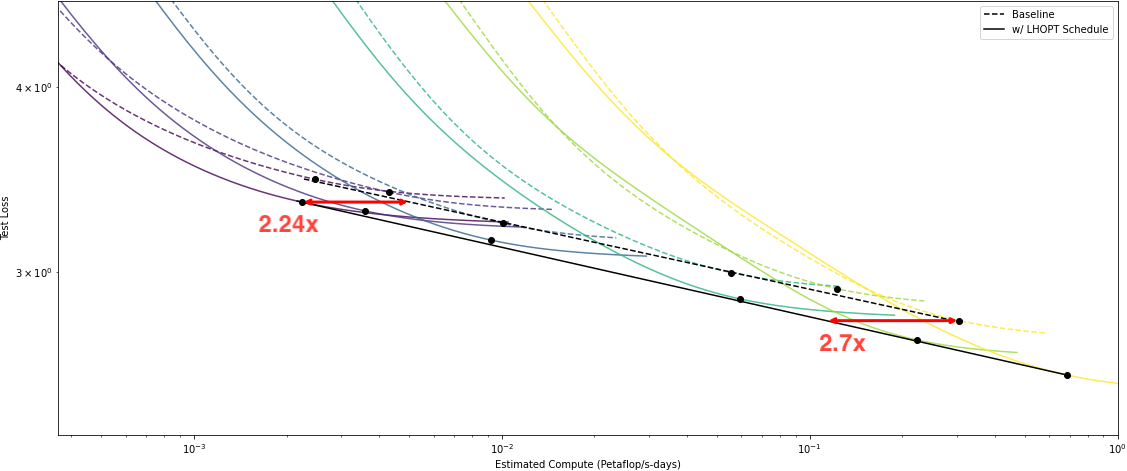}
    \caption{Test learning curves and scaling law fit of compute efficient frontier on a large well-tuned language modeling codebase. Our learned optimizers demonstrate consistent speedups $\geq$ 2x, with speedup increasing as model size does with no computational overhead. Dotted lines are baselines, full lines use a LHOPT hyperparameter schedule from a similar but smaller task.}
    \label{fig:lm_speedup}
\end{figure}

Our LHOPTs manage to generalize without tuning. They achieve approximately a 2x speedup on ImageNet~\citep{imagenet} compared to tuning both AdamW~\citep{adamw} learning rate and schedule. They outperform both baselines from MLPerf~\citep{mlperf} that are most different our training distribution: speech recognition and neural collaborative filtering. Finally, using a LHOPT's hyperparameter schedule from a smaller language modeling task, we apply it to a well-tuned language modeling codebase resulting in 2.5x speedups. In this scenario, despite the optimizer being trained on inner optimization problems that average only 170 GPU-seconds, it generalizes to all models that take up to 300 GPU-days (See Figure \ref{fig:lm_speedup}) - generalization of over 5 orders of magnitude.

We are not claiming that using a LHOPT is always better than default optimizers for neural network tasks, nor that LHOPTs would be the best for getting state of the art results. We see the primary contribution of this work as demonstrating that learned optimizers can generalize to real neural network tasks, and hope that it will inspire further efforts which deviate from the standard approaches and focus purely on generalization.

\section{Related Work}

Our work is most heavily inspired by a recent body of work that uses neural networks to learn parameter updates~\citep{l2l,scale_generalize,pathologies,luke2020}. Each subsequent refinement achieved large improvements in generalization capabilities, but even the most recent still had issues generalizing to slightly wider networks on the same task~\citep{luke2020}.

Another similar branch of work is on using hypergradients~\citep{maclaurin2015gradientbased,baydin2018online,grazzi2020iteration,moskovitz2020firstorder}, direct gradients from loss to hyperparameters, in order to optimize those hyperparameters on-the-fly. These methods have the advantage of not requiring training a policy ahead of time, but generally suffer from short-horizon bias~\citep{short_horizon_bias}.

Most similar to our work are those that use reinforcement learning to control learning rate~\citep{daniel2016learning,xu2017reinforcement,xu2019learning}. We interpret these models as learning rate-only versions of our LHOPT. Despite most of their results being on small scale datasets, the results of this paper indicate that their results could possibly generalize to large scale models.

\section{Learning Hyperparameter Updates}

Learning to update hyperparameters as opposed to parameters affords us 2 main axes of decoupling.

The first is our policy network architecture is decoupled from the details of the inner optimization task. For example, an optimizer that learns parameter updates would at least require access to gradient values, as well as possibly parameter values and other statistics. The problem with these statistics is that they may not be robust since they can assume very different values on out-of-distribution tasks, and thus result in undefined behavior. Instead, by letting the inner optimizer (e.g. AdamW) interact with these low level statistics, we have more flexibility in the types of features our policy uses for decision making and we can trade-off between giving more information and avoiding overfitting to idiosyncrasies of our training set.

The second is that we decouple the frequency at which our policy has to run. By definition, learned parameter optimizers have to run on every update, but LHOPTs can be run at any frequency, independent of the number of updates of the problem. This benefits both generalization and solves the problem of short-horizon bias~\citep{short_horizon_bias}: it allows us to both use the same number of steps for extremely large problems as that of smaller ones, and also optimize solely for final reward, as opposed to optimizing for reward after some fixed number of unrolled steps. This has the additional benefit of greatly decreasing the computational overhead of learned optimizers.

Given these benefits, we formulate the problem of learning hyperparameter optimizers as that of reinforcement learning, optimizing solely for final reward (Section \ref{reward_function}). We choose PPO~\citep{ppo} out of convenience. The LHOPT formulation allows us much flexibility, and we describe our more details for our choices of input features (Section \ref{feature_space}) and actions (Section \ref{action_space}), though we do not claim any design decision to be optimal and haven't tested alternatives unless otherwise specified.

\subsection{Problem Setup}

Our LHOPTs are designed to be applied to an inner task - a standard optimization problem, which uses an inner optimizer to perform updates every inner step (this would be the standard training step/update). Some predefined number of times throughout inner training, we also take an outer step where features from the inner task are summarized and passed into policy network $\pi$, which then outputs hyperparameter updates for the inner optimizer. These updates are applied and inner training resumes until the next outer step.

When training our learned optimizers, at the end of training each inner task, we map the final validation loss into a specified reward, which the PPO algorithm uses to update the weights of our policy network. Figure \ref{fig:diagram} shows a simplified view of the training setup.

At evaluation time, the LHOPT's behavior is defined by the learned weights of its policy network, as well as its predefined feature space and action space. 

\subsection{Feature Space}
\label{feature_space}

Our guiding principle for designing the feature space was to prevent the policy from inferring the inner optimization task. To ensure good generalization, we wanted every feature to look similar from toy problems all the way to massive scale models beyond which we could afford to train on. As such, we focused on "unitless features". 

Existing learned optimizers often use statistics such as the gradient, square of the gradient, loss values, and exponential moving averages (EMAs) of these statistics, which can be highly specific to the inner optimization task. In contrast, the majority of our features were transformations: log-ratios, cosine similarities, averages of booleans, or what we call CDF features.

Some examples of unitless features that we use are log-ratio between training and validation loss, log of the noise scale, cosine similarity between gradient and momentum tensors, how often the gradient is clipped, whether or not the loss is NaN, progress (how far along through inner training we are), an encoding of the previously sampled action, and CDF features of the validation loss. For a complete list of features of our final optimizers, see Appendix \ref{appendix:features}.

The motivations for CDF features are that (1) there are use-cases for which we would like to know relative values of a feature within an inner task and (2) ranking features should be invariant to details of the inner task. One example is to know if validation loss is plateauing without seeing its exact value. To achieve this, we calculate an estimate of that value's mean and variance (see Appendix \ref{appendix:integral_cdf} for how we do this in a robust way), and map that feature with a gaussian cumulative distribution function onto the interval [0,1]. Note that these features lose a lot of information: the first feature value is always 0.5, the second feature is 1 if larger than the first or 0 if smaller, etc.

One detail related to using PPO is that the value function $V$ can be given non-robust features~\citep{pinto2017asymmetric} in addition to the robust features given to the policy $\pi$.

As a final step of preprocessing, all input features are normalized by subtracting an exponential moving average and dividing by an exponential moving standard deviation (both with $\beta=0.999$). We additionally clip every input feature element-wise to [-2, 2] with the purpose of not being able to differentiate outliers.

\textbf{Limitations.} (1) One requirement that prevents plug-and-play usage of our LHOPTs is they require progress, training loss, and validation loss as an input from the user at evaluation time. Note that it is entirely possible to train a LHOPT with any or all of these features missing at the expense of performance. (2) As a result of making the validation loss observable by the policy, it is theoretical possible to overfit the validation dataset.. This can be solved with a separate validation set for this loss, but in practice, this doesn't appear to be an issue (Appendix \ref{appendix:overfit}). (3) Some features can be expensive to calculate on every inner step - for those features (specified in Appendix \ref{appendix:features}), we instead calculate them intermittently every 4 inner steps to reduce the overhead.

\subsection{Action Space}
\label{action_space}

For our actions, we limited our models to just two ways of updating hyperparameters: scaling (multiplying by a constant) for most hyperparameters, and logit shifting (applying the logit function, adding a constant, then applying a sigmoid) where we want to be careful when changing values near 0 and 1. By limiting our policy to relative actions and including randomness in the initial hyperparameters, models can never know the current values of hyperparameters and we make our model more robust by forcing them to react to the underlying inner task and not simply set hyperparameters to a specific value. Our hypothesis is that one set of hyperparameters on one task might behave very similarly to another set of hyperparameters on a different task.

We also had one additional class of actions that were not hyperparameter updates but fit in nicely within the existing framework: learning to restart from checkpoints. There are many motivations for such an action: (1) ideally learned optimizers would be able to handle all the task-specific tuning that a practitioner would have to do and restarting on divergence is one such tasks, (2) previous work has noted that SGD often works best with the highest possible stable learning rate~\citep{cyclical} and it may not be possible to determine that value without venturing into unstable territory, (3) sophisticated hyperparameter optimizations algorithms such as Population Based Training~\citep{pbt} could be learned from such a simple action, and finally (4) even if restarting was never used by a trained model, it could greatly help with exploration while training - to both decrease the length of credit assignment paths and also make it less punishing for models to sample suboptimal settings.

All actions were represented as discrete choices rather than real values, as this helps the stability of the RL training. In order to aid with exploration, we removed the identity action for many hyperparameters to encourage setting their values dynamically - in other words, by forcing the controller to always change the hyperparameter, we force it to be reactive to the inner problem instead of sticking with a safe initial value.

\textbf{Limitations.} (1) Despite the motivation, the policy could theoretically find the initial hyperparameters because most hyperparameters are clipped to minimum and maximum values. The optimizer could find when that clipping occurs, then restart to the beginning with that knowledge. We did not observe this behavior in our models, but it is possible that, as the techniques become more sophisticated, this design decision would need to be revisited. (2) In practice, our models did nothing close to the sophistication of PBT with restart actions, though they did use restarts effectively when divergence occurred, especially when given highly unstable initial hyperparameters. We hypothesize that the policy may be undertrained (see Section \ref{training_details}) and these sophisticated behaviors could emerge with larger-scale training.

\subsection{Reward Function}
\label{reward_function}

There are many desirable properties of a reward function: handling problems with very different losses, returning reward values that correspond to how difficult improvement was on a problem, being able to extrapolate beyond any baseline performance, and being computable on-the-fly to enable a potentially infinite inner problem distribution. Previous work mostly sacrifices the last two properties by computing many baselines and normalizing by the best one~\citep{luke2020}.

Instead, we use the following reward function: we first evaluate a baseline on the exact same problem, including all task specific values and initial weights. Then, we take all the points on the learning curve of that baseline and fit a power law curve to it. The result is a function where 0 corresponds to the worst loss, 1 to the best loss, and values greater than 1 take into account how difficult that problem is (i.e., we can extrapolate past the best loss of a baseline). While any baseline could be used here, we use an approach inspired by self-play~\citep{selfplay,alphago} and use the same learned optimizer policy that we were training but using an EMA of the weights, thus allowing our reward function to improve throughout outer training.

The main downside of naive application of this reward function is that it doubles the overhead of inner training. Instead, we choose to repeat each task four times to amortize the overhead of computing a baseline.

\subsection{Inner Optimizer}

While the LHOPT formulation would apply to any optimizer\footnote{See Appendix \ref{appendix:old_imagenet} for an older model which was used to set LR for both SGD and Adam.}, we choose to focus on our own Customizable Inner Adaptive Optimizer (CIAO)\footnote{Source code for our inner optimizer is available at \url{https://github.com/openai/LHOPT}}. The reason for this is that existing optimizers have been designed to work with mostly static hyperparameters and minimal hyperparameter tuning. By employing an inner optimizer designed to work with a LHOPT policy, hyperparameters turn from a weakness into a strength, and we can more easily create optimization techniques without worrying about how to tune them.

As such, our inner optimizer not only generalizes many existing standard and non-standard optimizers, we also introduce several novel techniques to improve optimization quality, including one to automate clipping gradients based on an exponentially weighted moving maximum of the gradient norm. Additionally, the default settings for our inner optimizer involve using the denominator of Adamax~\citep{adam} and the trust ratio of LAMB~\citep{lamb}. See Appendix \ref{appendix:inner_optimizer} for more inner optimizer details and Appendix \ref{appendix:why_lamb} for rationale on why our inner optimizers are use the LAMB trust ratio.

\textbf{Limitations.} (1) Despite the inner optimizer is closer to LAMB or Adamax than Adam, we still compare to Adam for most of the results. We choose to do so because of the prevalence of use of Adam as well as having a better understanding of how to tune learning rates and schedules for Adam. (2) Having CIAO automate more than Adam is desirable property from a usage perspective, but an undesirable one from the perspective of scientific comparison: it would be interesting future work to disentangle how much benefit comes from learning to optimize versus having more functionality. Section \ref{large_language_models} shows some evidence that majority of the benefit comes from learning to optimize.

\subsection{Training Details}
\label{training_details}

Inner tasks were a combination of MLPs, CNNs, RNNs, and Transformers (Appendix \ref{appendix:training_distribution} goes into more detail of the inner task distribution). Tasks had between 2 and 128 outer steps and had a median, mean, and max runtime of 149s, 170s, and 220s, respectively. Default settings of PPO were used, other than an maximum environment reuse of 4 (each rollout included in up to 4 batches for PPO training). Both the policy $\pi$ and value function $V$ used the same architecture, and our default architecture was a single hidden layer LSTM~\citep{lstm} with 256 hidden units. We trained all models for 16k PPO iterations resulting in less than 244k inner problems over the course of slightly under 512 GPU-days for our default model. Models were only optimized for final reward with discount factor $\gamma=1$, but we also performed reward shaping~\citep{shaping} using intermediate validation losses to aid with optimization.

\section{Results}

\textbf{Limitations.} One difficulty when evaluating schedules (dynamic or otherwise) is that they optimize for a chosen completion time: a single run only provides an approximate upper or lower bound of the speedup and cannot give you an exact measurement of a speedup, because the latter would involve changing that number of updates. Measuring precise speedups would require an iterative search over many different end times for a schedule. As such, for most of our evaluations, we only compare to with the same and half as many iterations as baselines to verify $\geq 1$x and $\geq 2$x speedups, respectively.

\subsection{On a Distribution of Small Problems}
\label{traindist}

Our main development evaluation metric compares our trained models on almost 4000 tasks similar to those in the training distribution to a set of 35 baselines using a very simple metric: the fraction of tasks where the model gets a $\geq$ 2x speedup. We compute this metric by running each baseline with twice as many updates, and compare the model's performance on every task to that of every baseline on that task. The baselines are all AdamW-based and combinations of 5 learning rates ($1e^{-4}, 3e^{-4}, 1e^{-3}, 3e^{-3}, 1e^{-2}$) and 7 commonly used schedules (constant, multi-step, linear decay, quadratic decay, exponential decay, cosine~\citep{cosine} to 0, cosine to 0.1 of original LR). Early on in development, we looked at the fraction of tasks where the model gets a $\geq$ 1x speedup (i.e., fraction of tasks where the model beats baselines), but the models quickly dominated all the baselines at that threshold.

Table \ref{traindist_table} shows overall scores for several models, as well as scores for different subsets of tasks. While the metrics are meaningless for real world performance, they do show the relative strength of learned optimizers, as well as some amount of generalization: the models generalization to Transformers~\citep{transformer} on algorithmic tasks despite never being trained on them.

\textbf{Limitations.} (1) These tasks have some small similarity with those in the training distribution. Due to the large amount of randomness of our training distribution, it would be possible but very unlikely that any of the exact settings of our test set were sampled during training. (2) Some of the 35 baselines might be much weaker than one would use on a real problem - as such this metric almost certainly overestimates the rate at which our model would get $\geq$ 2x speedups even with perfect generalization. (3) This metric ignores how much better or worse the models are than baselines. (4) It is quite possible that a smaller subset of tasks provides the same information and this metric may be unnecessarily computationally expensive, costing approximately 64 V100-days.

\begin{table}
  \caption{Fraction of test tasks where the model gets a $\geq$ 2x speedup}
  \label{traindist_table}
    \centering
    \begin{tabular}{lrrrr}
    
        \toprule
        {} &  Default &  OnlyMNISTs &  OnlyNQM &  Large \\
        \midrule
        Overall            &    0.691 &        0.631 &     0.294 &  0.725 \\
        \midrule
        MNISTs + MLP       &    0.443 &        0.479 &     0.073 &  0.457 \\
        MNISTs + CNN       &    0.777 &        0.829 &     0.095 &  0.811 \\
        Algo + RNN         &    0.817 &        0.637 &     0.469 &  0.893 \\
        Text + RNN         &    0.827 &        0.719 &     0.478 &  0.853 \\
        Algo + Transformer &    0.572 &        0.580 &     0.340 &  0.599 \\
        Text + Transformer &    0.713 &        0.543 &     0.309 &  0.738 \\
        \bottomrule

    \end{tabular}
\end{table}

\subsection{GPT-2 on 1 Epoch of WikiText-103}
\label{huggingface}

We test our model's ability to generalize to large language models by using it to train a 760M parameter GPT2-Large~\citep{gpt2} model on a single epoch of WikiText-103~\citep{wikitext103} using a standard HuggingFace~\citep{huggingface} setup. Because even the largest language modeling tasks are trained for less than an epoch~\citep{gpt3}, we choose to train for only a single epoch to evaluate performance in an underfitting regime. For this task, AdamW was used as a baseline and grid search was used over both learning rate and schedule (over constant, linear, and cosine). The LHOPT was not tuned at all. Table \ref{huggingface_table} shows the relative performance with our LHOPT greatly outperforming at 1x time and almost showing a 2x speedup. 

\begin{figure}[h]
    \centering
    \includegraphics[scale=0.5]{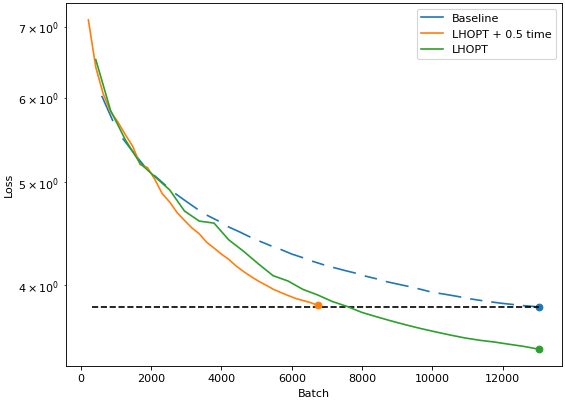}
    \caption{Performance of learned optimizers on optimizing 1 epoch of GPT2-Large on WikiText-103. Our learned optimizers get almost 2x speedups on this task despite being over 2 magnitudes larger than training tasks.}
    \label{fig:huggingface}
\end{figure}

Figure \ref{fig:huggingface} shows the learning curves for the LHOPTs and best baseline. An interesting observation that we will see repeated throughout the paper is that despite being capable of achieving a lower loss earlier, the chosen hyperparameters tend to underperform the best possible loss for that compute, presumably to achieve a better loss later. It's unclear how necessary it is trade-off early performance for later, but many successful hand-made schedules tend to do this: multi-step schedules tend to stay at the same learning rate long after they've hit a plateau and cosine schedules tend to decay their learning rates much less aggressively than other commonly used schedules.  

\textbf{Limitations.} Despite the LHOPT not being tuned at all, this evaluation was used a decent amount during development, and it's possible that some implicit overfitting has occurred.

\begin{table}
  \caption{Language modeling performance on WikiText-103}
  \label{huggingface_table}
  \centering
  \begin{tabular}{ll}
    \toprule
    Model & Test Perplexity \\
    \midrule
    GPT2-Large & 45.6 \\ 
    + LHOPT + 0.5 time & 46.1 \\
    + LHOPT & $\mathbf{32.5}$ \\
    + LHOPT + OM + 0.5 time & 63.4 \\
    + LHOPT + OM & 40.9 \\
    \bottomrule
  \end{tabular}
\end{table}

\subsection{ResNets on ImageNet}
\label{imagenet}

We test our model's ability to generalize to novel architectures by training ResNets~\citep{resnet} on ImageNet. Table \ref{imagenet_table} shows performance relative to an AdamW baseline tuned with grid search over both learning rate and schedule (over constant, cosine, and multi-step) and Table \ref{old_imagenet_table} also includes a tuned SGD baseline. Performance is reported on half of the validation set, where the other half is used for either tuning for the baseline or as an input validation loss for the LHOPT. Despite performing worse than SGD, our LHOPTs greatly outperform baselines resulting in an approximately 2x speedup despite having no ResNets on our training set.

\begin{figure}[h]
    \centering
    \includegraphics[scale=0.5]{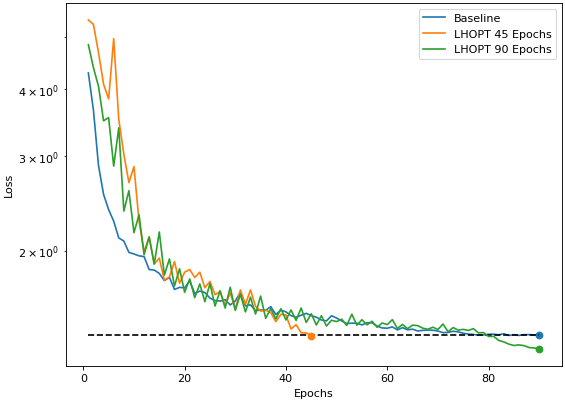}
    \caption{Test learning curve of LHOPTs and a tuned AdamW optimizing ResNet18 on ImageNet. Learned optimizers outperform the baseline adaptive optimizer despite never encountering residual architectures in its training set.}
    \label{fig:imagenet_resnet18}
\end{figure}

Figure \ref{fig:imagenet_resnet18} shows learning curves for ResNet18 models, similarly showing learned have the capacity to achieve lower loss earlier in training.

\textbf{Limitations.} (1) Note that we choose to compare to AdamW instead of SGD because we constrain our optimizer to be adaptive. While past work has noted that a more general optimizer should be able to outperform one they can approximate~\citep{empirical_comparisons}, the bounds on our inner optimizer's hyperparameter values prevent it from doing so (Appendix \ref{appendix:inner_optimizer}). Appendix \ref{appendix:old_imagenet} also contains a comparison to LHOPTs using SGD as an inner optimizer. (2) AdamW is a weak baseline for ResNets on ImageNet.

\begin{table}
  \caption{ImageNet performance of adaptive optimizers + SGD baselines for reference}
  \label{imagenet_table}
  \centering
  \begin{tabular}{lllll}
    \toprule
    Model & Epochs & Acc@1 & Acc@5 & Test Loss \\
    \midrule
    Resnet18 + AdamW & 90 & 67.32 & 87.52 & 1.39 \\
    Resnet18 + LHOPT & 45 & 67.24 & 87.54 & 1.39 \\
    Resnet18 + LHOPT & 90 & $\mathbf{68.89}$ & 88.43 & $\mathbf{1.31}$ \\
    Resnet18 + LHOPT + OM & 45 & 66.54 & 87.09 & 1.41 \\
    Resnet18 + LHOPT + OM & 90 & 68.34 & $\mathbf{88.54}$ & $\mathbf{1.31}$ \\
    \midrule 
    Resnet50 + AdamW & 90 & 71.42 & 89.66 & 1.35 \\
    Resnet50 + LHOPT & 45 & 72.88 & 91.17 & 1.14 \\
    Resnet50 + LHOPT & 90 & $\mathbf{73.52}$ & $\mathbf{91.38}$ & $\mathbf{1.07}$ \\
    \midrule
    Resnet18 + SGD & 90 & $\mathbf{69.10}$ & $\mathbf{88.62}$ & $\mathbf{1.26}$ \\
    Resnet50 + SGD & 90 & $\mathbf{74.66}$ & $\mathbf{92.09}$ & $\mathbf{1.01}$ \\
    \bottomrule
  \end{tabular}
\end{table}

\subsection{Additional Out-of-distribution Evaluations}

To additionally test the limits of generalization of our learned optimizers, we also evaluated them on 2 tasks in the MLPerf suite: the NeuMF model~\citep{ncf} for recommendation and Deep Speech 2~\citep{ds2} for speech recognition. These 2 tasks were chosen as the most different from tasks in our training set which had an available PyTorch~\citep{pytorch} implementation from the MLPerf training reference implementations.

\subsubsection{Neural Collaborative Filtering}
\label{ncf}

We trained both the Generalized Matrix Factorization (GMF) and Neural Matrix Factorization (NeuMF) models from scratch on the MovieLens 1M Dataset. Table \ref{ncf_table} shows the results for both models. Interestingly, the LHOPT underperformed the baseline for the GMF models, but beat the baseline for the NeuMF model. This is the only instance we have found where a LHOPT underperforms a baseline. We hypothesize that this is due to our model's behavior specializing in optimizing deep networks and the GMF model being shallow (no hidden nonlinearities). As an additional surprise, the LHOPT worked without any tuning on the NeuMF model, despite their default settings failing to train and their README recommending tuning weight decay in order to get it working.

\begin{figure}[h]
    \begin{subfigure}{.5\textwidth}
        \includegraphics[scale=0.4]{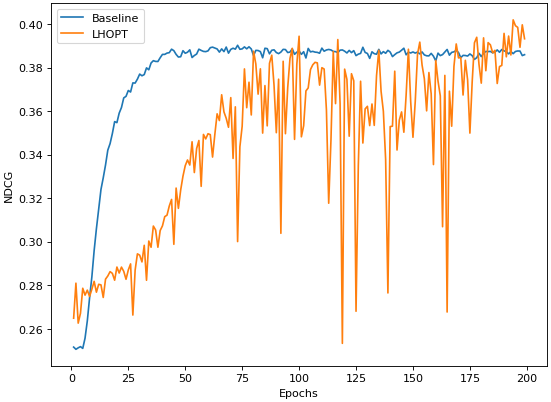}
        \caption{}
        \label{fig:ncf_ncdg}
    \end{subfigure}%
    \begin{subfigure}{.5\textwidth}
        \includegraphics[scale=0.4]{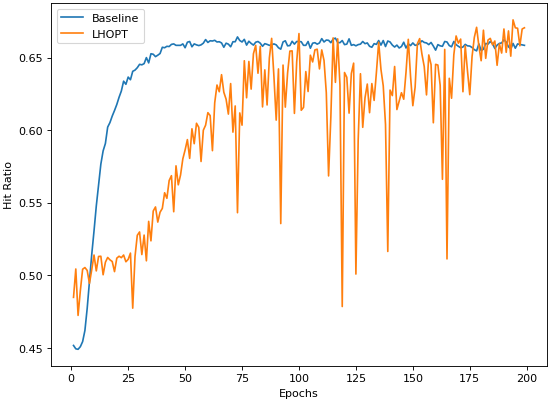}
        \caption{}
        \label{fig:ncf_hr}
    \end{subfigure}
    \caption{Comparison on NeuMF model of (\ref{fig:ncf_ncdg}) NCDG and (\ref{fig:ncf_hr}) hit ratio. These demonstrate generalization to out of distribution architecture, input modality, and loss function without tuning. Hyperparameters are updated 200 times (every epoch).}
    \label{fig:ncf}
\end{figure}

Figure \ref{fig:ncf} shows validation metrics over time where the performance plots show the learned optimizer training to look comparatively less stable. One potential benefit of using LHOPTs might be in being able to train with less stable dynamics without divergence.

Because their codebase doesn't expose a evaluation loss similar to their training loss, we simply pass in negative NCDG as our validation loss to the LHOPT. The fact that this still works demonstrates the flexibility of our unitless features.

\textbf{Limitations.} (1) Despite having to do minimal tuning of weight decay to get the NeuMF model, we only compare to the baseline learning rates and do no tuning of learning rate or schedule of our own. We assume that being part of MLPerf's training suite, the model has already been tuned, but we are unsure of how thoroughly. (2) The presented metrics are on the validation set, which the LHOPT could theoretically overfit to.

\begin{table}
  \caption{NCF Performance. Larger is better for both metrics.}
  \label{ncf_table}
  \centering
  \begin{tabular}{lll}
    \toprule
    Model & NCDG & Hit Ratio \\
    \midrule
    GMF & $\mathbf{0.3677}$ & $\mathbf{0.6397}$ \\ 
    GMF + LHOPT & 0.3134 & 0.5553 \\
    \midrule 
    NeuMF & 0.3859 & 0.6584 \\
    NeuMF + LHOPT & $\mathbf{0.3932}$ & $\mathbf{0.6705}$ \\ 
    \bottomrule
  \end{tabular}
\end{table}

\subsubsection{Speech Recognition}

\begin{figure}[h]
    \centering
    \includegraphics[scale=0.4]{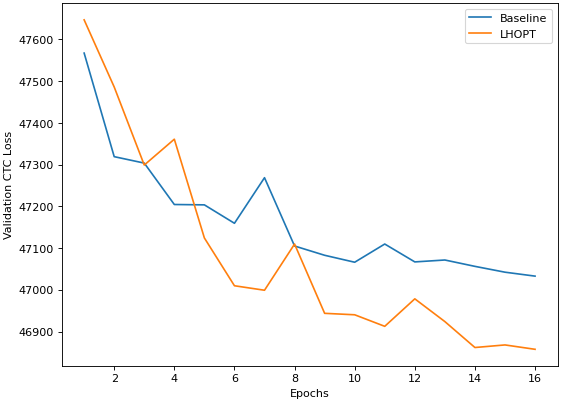}
    \caption{Loss curves for Deep Speech 2 on LibriSpeech. Similar to Figure \ref{fig:ncf}, this plot demonstrates generalization of our learned optimizers to inner tasks with out of distribution size, input modality, loss function, and fp16 training without tuning. Hyperparameters are updated 16 times (every epoch) demonstrating robustness to outer step frequency.}
    \label{fig:deepspeech2}
\end{figure}

Figure \ref{fig:deepspeech2} shows our LHOPT outperforming the baseline MLPerf optimization. This demonstrates generalization to a modality that our models have never been exposed to as well as working with half-precision floating point.

\textbf{Limitations.} (1) Similarly to NCF above, we only compare to a single baseline and do no tuning of our own. (2) The presented metrics are on the validation set, which the LHOPT could theoretically overfit to.

\subsection{Fixed Schedules for Large Language Models}
\label{large_language_models}

To explore another axis of generalization, we tested our models on a well-tuned language modeling codebase~\citep{gpt3}. To avoid the software complexity of making our optimizer work in a distributed setting, we instead investigated the effectiveness of transferring a fixed hyperparameter schedule created by our optimizer on a similar task at a smaller scale. Experiments on WikiText-103 seemed to indicate that majority of the benefit of LHOPTs could be had by running the optimizer once on a similar problem and using the resulting schedule retains the majority of performance.

We thus retrained a new LHOPT that uses an inner optimizer with the exact same hyperparameters as those available in the distributed codebase instead of CIAO with only 4 hyperparameters to control: learning rate, $\epsilon$, $\beta_2$, and weight decay. This model had slightly degraded performance compared to our defaults on the metric from Section \ref{traindist}: 0.673 for this LHOPT versus 0.691 for the default. We ran this model on the same setting as in Section \ref{huggingface}, achieving a slightly worse perplexity of 37.7, and used the resulting schedule for the remainder of this section. See Appendix \ref{appendix:fixed_schedule} for this schedule.

We then trained a range of model sizes to compute scaling laws~\citep{scaling_laws} for both baselines and models trained with the LHOPT schedule and present the results on Figure \ref{fig:lm_speedup}. The LHOPT schedule demonstrates consistent speedup over baselines with a slightly steeper slope. We can estimate what a constant speedup would be for this range of points by assume their scaling law slopes are equal and from this calculate a 2.5x speedup. To take the change in slope into account as well, we extrapolate the curves to 175 billion parameters (same size as GPT-3) and at that size, the estimated speedup would be 3.6x.

Note that this result is despite the codebase doing multiple optimization techniques that our LHOPT would have no way of taking into account: gradient clipping to a fixed value and gradually increasing batch size.

\textbf{Limitations.} In order to run the schedule on this task, one important hyperparameter is how far we set the end of the schedule. Since we are comparing curves of the compute efficient frontier, we cannot simply set it to a single point that we care about. Instead, we choose to set that end point to twice the number of updates as that of the efficient point of the baseline. We found performance to be robust to how this end point was set, but given how our models tend to not optimize for performance before their end point, these frontiers almost certainly underestimate their performance. 

\subsection{How Important is Task Diversity?}
\label{task_diversity}

Previous work has emphasized the importance of task diversity~\citep{luke2020} in the context of learned parameter optimizers, and we would like to explore its importance for our generalization-first approach. To do so, we train additional LHOPTs on only MNIST-like datasets (OM) and on only NQMs~\citep{nqm}. Table \ref{traindist_table} shows that training on NQMs only causes a large drop in performance but training on only MNISTs\footnote{Note that MNISTs refer to the original MNIST~\citep{mnist}, KMNIST~\citep{kmnist}, and Fashion-MNIST~\citep{fashionmnist}.} recovers majority of that performance. Notably, we see the latter model generalize well to all other tasks. We test the OM model on the same setups as Sections \ref{huggingface} and \ref{imagenet} and include their results in Tables \ref{huggingface_table} and \ref{imagenet_table} respectively. On the language modeling task, the OM model does worse than the LHOPT trained on the full training set, but still much better than the baseline despite never being trained on anything other than 28x28 images and never being exposed to an attention mechanism, and on ImageNet, the OM does somewhat comparably to having the full dataset.

We believe that these results imply that for LHOPTs constructed in the way described, task diversity is not as important as it is for previous work, though these results are definitely not conclusive. Another possibility is that the flexibility in our reward function allowed us to include more randomness in task specification, and thus provided enough diversity from just those 3 MNIST-like datasets: there were over 50 billion different task settings we were sampling from, not including sampling initial hyperparameters.

\subsection{Computational Overhead}

Computational overhead for running the learned optimizer depends on the size for the inner problem. Running the policy network involves a constant amount of calculation to run the LSTM network for a fixed number of outer steps which would be insignificant for most real world problems, but the majority of computational overhead comes from computing features for the policy, which using our default setup would be proportional to the number of parameters of the inner problem. This overhead ranges from less than 2\% on ResNets to 20\% on Transformers, which could be reduced by recalculating inner-features less frequently - by default, we only update inner-feature statistics every 4 optimizer steps or simply using fixed schedules.

The memory overhead of our default setup is similar to Adam and other adaptive methods: 2 accumulators per parameter. Their is additional constant memory cost to run the LSTM controller and keep track of the features we compute, but this is negligible for practical problems. This is significantly lower than existing learned optimizers which require many times the memory of Adam~\citep{luke2020} due to having a per-parameter neural network running.

\section{Conclusion}

We present a novel approach for learning optimizers that prioritizes generalization above all. We demonstrate successful generalization over a variety of tasks very distinct from the training distribution including 2x speedups on ImageNet, 2.5x speedups on large scale language modeling, and outperforming tuned Adam baselines on a large set of practical neural network task.

The failures of the OnlyNQM LHOPT to generalize in Section \ref{task_diversity} as well as the GMF task in Section \ref{ncf} being the only task we could find where our LHOPT underperformed baselines lead us to believe that there is some commonality in optimizing specifically deep networks that our optimizers exploit. We believe that these results show a hopeful picture for learning optimizers that generalize.

\section{Future Work}

We view this work as preliminary work on getting learned optimizers to generalize rather than a thorough analysis of all the components of our system. Many ablations could be done to better understand how impactful each of the individual design decisions were, such as applying our reward model and inner task distribution to existing learned parameter optimizers, viewing LHOPT performance with and without the constraints of unitless features and/or relative hyperparameter updates, analyzing the importance of initial hyperparameter randomness, and controlling for the effect of inner optimizer.

While this work focuses on a more standard training setup, we believe it would be incredibly useful to practitioners to extend it to other setups including but not limited to GANs, reinforcement learning, fine-tuning, and multi-loss training.

LHOPTs could also be used in combination with learned parameter optimizers: an outer-most LHOPT could control hyperparameters related to generalization and short horizon bias, which would allow learned parameter optimizers to greatly increase the per-step learning capacity, getting the best of both worlds.

We also believe that LHOPTs could greatly assist other subfields of research: there have been many promising optimization primitives emerging that so far have been unsuccessful but perhaps just need the right hyperparameters to get working such as SVRG~\citep{svrg}, hypergradients~\citep{maclaurin2015gradientbased}, and second-order methods~\citep{shampoo1,shampoo2,kfac}. In fact, there could be many opportunities to improve performance using dynamic hyperparameter-controlled architectures~\citep{gradnets}: hyperparameters need not be a bad thing if the user doesn't have to set them.

\section{Acknowledgements}

We would like to thank Luke Metz, Sander Dieleman, Alec Radford, Arthur Caillau, and Ryan Lowe for discussion related to this work and everyone at OpenAI, especially those with unused compute capacity that is made available for low priority jobs. We would like to especially thank Luke Metz, whose discussions and previous work in the field of learned optimizers inspired our generalization-first direction.

\bibliography{main}

\begin{thebibliography}{54}
\providecommand{\natexlab}[1]{#1}
\providecommand{\url}[1]{\texttt{#1}}
\expandafter\ifx\csname urlstyle\endcsname\relax
  \providecommand{\doi}[1]{doi: #1}\else
  \providecommand{\doi}{doi: \begingroup \urlstyle{rm}\Url}\fi

\bibitem[Almeida and Sauder(2015)]{gradnets}
D.~Almeida and N.~Sauder.
\newblock Gradnets: Dynamic interpolation between neural architectures.
\newblock \emph{arXiv preprint arXiv:1511.06827}, 2015.

\bibitem[Amodei et~al.(2016)Amodei, Ananthanarayanan, Anubhai, Bai, Battenberg,
  Case, Casper, Catanzaro, Cheng, Chen, et~al.]{ds2}
D.~Amodei, S.~Ananthanarayanan, R.~Anubhai, J.~Bai, E.~Battenberg, C.~Case,
  J.~Casper, B.~Catanzaro, Q.~Cheng, G.~Chen, et~al.
\newblock Deep speech 2: End-to-end speech recognition in english and mandarin.
\newblock In \emph{International conference on machine learning}, pages
  173--182. PMLR, 2016.

\bibitem[Andrychowicz et~al.(2016)Andrychowicz, Denil, Gomez, Hoffman, Pfau,
  Schaul, Shillingford, and De~Freitas]{l2l}
M.~Andrychowicz, M.~Denil, S.~Gomez, M.~W. Hoffman, D.~Pfau, T.~Schaul,
  B.~Shillingford, and N.~De~Freitas.
\newblock Learning to learn by gradient descent by gradient descent.
\newblock \emph{arXiv preprint arXiv:1606.04474}, 2016.

\bibitem[Anil et~al.(2020)Anil, Gupta, Koren, Regan, and Singer]{shampoo2}
R.~Anil, V.~Gupta, T.~Koren, K.~Regan, and Y.~Singer.
\newblock Second order optimization made practical.
\newblock \emph{arXiv preprint arXiv:2002.09018}, 2020.

\bibitem[Babanezhad et~al.(2015)Babanezhad, Ahmed, Virani, Schmidt, Konečný,
  and Sallinen]{svrg}
R.~Babanezhad, M.~O. Ahmed, A.~Virani, M.~Schmidt, J.~Konečný, and
  S.~Sallinen.
\newblock Stop wasting my gradients: Practical svrg, 2015.

\bibitem[Baydin et~al.(2018)Baydin, Cornish, Rubio, Schmidt, and
  Wood]{baydin2018online}
A.~G. Baydin, R.~Cornish, D.~M. Rubio, M.~Schmidt, and F.~Wood.
\newblock Online learning rate adaptation with hypergradient descent, 2018.

\bibitem[Bengio(2012)]{bengio2012practical}
Y.~Bengio.
\newblock Practical recommendations for gradient-based training of deep
  architectures, 2012.

\bibitem[Brown et~al.(2020)Brown, Mann, Ryder, Subbiah, Kaplan, Dhariwal,
  Neelakantan, Shyam, Sastry, Askell, Agarwal, Herbert-Voss, Krueger, Henighan,
  Child, Ramesh, Ziegler, Wu, Winter, Hesse, Chen, Sigler, Litwin, Gray, Chess,
  Clark, Berner, McCandlish, Radford, Sutskever, and Amodei]{gpt3}
T.~B. Brown, B.~Mann, N.~Ryder, M.~Subbiah, J.~Kaplan, P.~Dhariwal,
  A.~Neelakantan, P.~Shyam, G.~Sastry, A.~Askell, S.~Agarwal, A.~Herbert-Voss,
  G.~Krueger, T.~Henighan, R.~Child, A.~Ramesh, D.~M. Ziegler, J.~Wu,
  C.~Winter, C.~Hesse, M.~Chen, E.~Sigler, M.~Litwin, S.~Gray, B.~Chess,
  J.~Clark, C.~Berner, S.~McCandlish, A.~Radford, I.~Sutskever, and D.~Amodei.
\newblock Language models are few-shot learners, 2020.

\bibitem[Cho et~al.(2014)Cho, van Merrienboer, Gulcehre, Bahdanau, Bougares,
  Schwenk, and Bengio]{gru}
K.~Cho, B.~van Merrienboer, C.~Gulcehre, D.~Bahdanau, F.~Bougares, H.~Schwenk,
  and Y.~Bengio.
\newblock Learning phrase representations using rnn encoder-decoder for
  statistical machine translation, 2014.

\bibitem[Choi et~al.(2020)Choi, Shallue, Nado, Lee, Maddison, and
  Dahl]{empirical_comparisons}
D.~Choi, C.~J. Shallue, Z.~Nado, J.~Lee, C.~J. Maddison, and G.~E. Dahl.
\newblock On empirical comparisons of optimizers for deep learning, 2020.

\bibitem[Clanuwat et~al.(2018)Clanuwat, Bober-Irizar, Kitamoto, Lamb, Yamamoto,
  and Ha]{kmnist}
T.~Clanuwat, M.~Bober-Irizar, A.~Kitamoto, A.~Lamb, K.~Yamamoto, and D.~Ha.
\newblock Deep learning for classical japanese literature.
\newblock \emph{arXiv preprint arXiv:1812.01718}, 2018.

\bibitem[Daniel et~al.(2016)Daniel, Taylor, and Nowozin]{daniel2016learning}
C.~Daniel, J.~Taylor, and S.~Nowozin.
\newblock Learning step size controllers for robust neural network training.
\newblock In \emph{Thirtieth AAAI Conference on Artificial Intelligence}, 2016.

\bibitem[Deng et~al.(2009)Deng, Dong, Socher, Li, Li, and Fei-Fei]{imagenet}
J.~Deng, W.~Dong, R.~Socher, L.-J. Li, K.~Li, and L.~Fei-Fei.
\newblock Imagenet: A large-scale hierarchical image database.
\newblock In \emph{2009 IEEE conference on computer vision and pattern
  recognition}, pages 248--255. Ieee, 2009.

\bibitem[Gers et~al.(1999)Gers, Schmidhuber, and Cummins]{lstm}
F.~A. Gers, J.~Schmidhuber, and F.~Cummins.
\newblock Learning to forget: Continual prediction with lstm.
\newblock 1999.

\bibitem[Glorot and Bengio(2010)]{glorot}
X.~Glorot and Y.~Bengio.
\newblock Understanding the difficulty of training deep feedforward neural
  networks.
\newblock In \emph{Proceedings of the thirteenth international conference on
  artificial intelligence and statistics}, pages 249--256. JMLR Workshop and
  Conference Proceedings, 2010.

\bibitem[Grazzi et~al.(2020)Grazzi, Franceschi, Pontil, and
  Salzo]{grazzi2020iteration}
R.~Grazzi, L.~Franceschi, M.~Pontil, and S.~Salzo.
\newblock On the iteration complexity of hypergradient computation, 2020.

\bibitem[Gupta et~al.(2018)Gupta, Koren, and Singer]{shampoo1}
V.~Gupta, T.~Koren, and Y.~Singer.
\newblock Shampoo: Preconditioned stochastic tensor optimization.
\newblock In \emph{International Conference on Machine Learning}, pages
  1842--1850. PMLR, 2018.

\bibitem[He et~al.(2015)He, Zhang, Ren, and Sun]{resnet}
K.~He, X.~Zhang, S.~Ren, and J.~Sun.
\newblock Deep residual learning for image recognition, 2015.

\bibitem[He et~al.(2017)He, Liao, Zhang, Nie, Hu, and Chua]{ncf}
X.~He, L.~Liao, H.~Zhang, L.~Nie, X.~Hu, and T.-S. Chua.
\newblock Neural collaborative filtering.
\newblock In \emph{Proceedings of the 26th international conference on world
  wide web}, pages 173--182, 2017.

\bibitem[Jaderberg et~al.(2017)Jaderberg, Dalibard, Osindero, Czarnecki,
  Donahue, Razavi, Vinyals, Green, Dunning, Simonyan, Fernando, and
  Kavukcuoglu]{pbt}
M.~Jaderberg, V.~Dalibard, S.~Osindero, W.~M. Czarnecki, J.~Donahue, A.~Razavi,
  O.~Vinyals, T.~Green, I.~Dunning, K.~Simonyan, C.~Fernando, and
  K.~Kavukcuoglu.
\newblock Population based training of neural networks, 2017.

\bibitem[Kaplan et~al.(2020)Kaplan, McCandlish, Henighan, Brown, Chess, Child,
  Gray, Radford, Wu, and Amodei]{scaling_laws}
J.~Kaplan, S.~McCandlish, T.~Henighan, T.~B. Brown, B.~Chess, R.~Child,
  S.~Gray, A.~Radford, J.~Wu, and D.~Amodei.
\newblock Scaling laws for neural language models, 2020.

\bibitem[Kingma and Ba(2014)]{adam}
D.~P. Kingma and J.~Ba.
\newblock Adam: A method for stochastic optimization.
\newblock \emph{arXiv preprint arXiv:1412.6980}, 2014.

\bibitem[Leclerc and Madry(2020)]{2regimes}
G.~Leclerc and A.~Madry.
\newblock The two regimes of deep network training, 2020.

\bibitem[LeCun et~al.(1998)LeCun, Bottou, Bengio, and Haffner]{mnist}
Y.~LeCun, L.~Bottou, Y.~Bengio, and P.~Haffner.
\newblock Gradient-based learning applied to document recognition.
\newblock \emph{Proceedings of the IEEE}, 86\penalty0 (11):\penalty0
  2278--2324, 1998.

\bibitem[Loshchilov and Hutter(2016)]{cosine}
I.~Loshchilov and F.~Hutter.
\newblock Sgdr: Stochastic gradient descent with warm restarts.
\newblock \emph{arXiv preprint arXiv:1608.03983}, 2016.

\bibitem[Loshchilov and Hutter(2018)]{adamw}
I.~Loshchilov and F.~Hutter.
\newblock Fixing weight decay regularization in adam.
\newblock 2018.

\bibitem[Maclaurin et~al.(2015)Maclaurin, Duvenaud, and
  Adams]{maclaurin2015gradientbased}
D.~Maclaurin, D.~Duvenaud, and R.~P. Adams.
\newblock Gradient-based hyperparameter optimization through reversible
  learning, 2015.

\bibitem[Martens and Grosse(2020)]{kfac}
J.~Martens and R.~Grosse.
\newblock Optimizing neural networks with kronecker-factored approximate
  curvature, 2020.

\bibitem[Mattson et~al.(2019)Mattson, Cheng, Coleman, Diamos, Micikevicius,
  Patterson, Tang, Wei, Bailis, Bittorf, et~al.]{mlperf}
P.~Mattson, C.~Cheng, C.~Coleman, G.~Diamos, P.~Micikevicius, D.~Patterson,
  H.~Tang, G.-Y. Wei, P.~Bailis, V.~Bittorf, et~al.
\newblock Mlperf training benchmark.
\newblock \emph{arXiv preprint arXiv:1910.01500}, 2019.

\bibitem[McCandlish et~al.(2018)McCandlish, Kaplan, Amodei, and
  Team]{mccandlish2018empirical}
S.~McCandlish, J.~Kaplan, D.~Amodei, and O.~D. Team.
\newblock An empirical model of large-batch training.
\newblock \emph{arXiv preprint arXiv:1812.06162}, 2018.

\bibitem[Merity(2019)]{sharnn}
S.~Merity.
\newblock Single headed attention rnn: Stop thinking with your head, 2019.

\bibitem[Merity et~al.(2016)Merity, Xiong, Bradbury, and Socher]{wikitext103}
S.~Merity, C.~Xiong, J.~Bradbury, and R.~Socher.
\newblock Pointer sentinel mixture models, 2016.

\bibitem[Metz et~al.(2019)Metz, Maheswaranathan, Nixon, Freeman, and
  Sohl-Dickstein]{pathologies}
L.~Metz, N.~Maheswaranathan, J.~Nixon, D.~Freeman, and J.~Sohl-Dickstein.
\newblock Understanding and correcting pathologies in the training of learned
  optimizers.
\newblock In \emph{International Conference on Machine Learning}, pages
  4556--4565. PMLR, 2019.

\bibitem[Metz et~al.(2020)Metz, Maheswaranathan, Freeman, Poole, and
  Sohl-Dickstein]{luke2020}
L.~Metz, N.~Maheswaranathan, C.~D. Freeman, B.~Poole, and J.~Sohl-Dickstein.
\newblock Tasks, stability, architecture, and compute: Training more effective
  learned optimizers, and using them to train themselves, 2020.

\bibitem[Moskovitz et~al.(2020)Moskovitz, Wang, Lan, Kapoor, Miconi, Yosinski,
  and Rawal]{moskovitz2020firstorder}
T.~Moskovitz, R.~Wang, J.~Lan, S.~Kapoor, T.~Miconi, J.~Yosinski, and A.~Rawal.
\newblock First-order preconditioning via hypergradient descent, 2020.

\bibitem[Ng et~al.(1999)Ng, Harada, and Russell]{shaping}
A.~Y. Ng, D.~Harada, and S.~Russell.
\newblock Policy invariance under reward transformations: Theory and
  application to reward shaping.
\newblock In \emph{Icml}, volume~99, pages 278--287, 1999.

\bibitem[OpenAI et~al.(2019)OpenAI, :, Berner, Brockman, Chan, Cheung, Dębiak,
  Dennison, Farhi, Fischer, Hashme, Hesse, Józefowicz, Gray, Olsson, Pachocki,
  Petrov, d.~O.~Pinto, Raiman, Salimans, Schlatter, Schneider, Sidor,
  Sutskever, Tang, Wolski, and Zhang]{dota2}
OpenAI, :, C.~Berner, G.~Brockman, B.~Chan, V.~Cheung, P.~Dębiak, C.~Dennison,
  D.~Farhi, Q.~Fischer, S.~Hashme, C.~Hesse, R.~Józefowicz, S.~Gray,
  C.~Olsson, J.~Pachocki, M.~Petrov, H.~P. d.~O.~Pinto, J.~Raiman, T.~Salimans,
  J.~Schlatter, J.~Schneider, S.~Sidor, I.~Sutskever, J.~Tang, F.~Wolski, and
  S.~Zhang.
\newblock Dota 2 with large scale deep reinforcement learning, 2019.

\bibitem[Paszke et~al.(2019)Paszke, Gross, Massa, Lerer, Bradbury, Chanan,
  Killeen, Lin, Gimelshein, Antiga, et~al.]{pytorch}
A.~Paszke, S.~Gross, F.~Massa, A.~Lerer, J.~Bradbury, G.~Chanan, T.~Killeen,
  Z.~Lin, N.~Gimelshein, L.~Antiga, et~al.
\newblock Pytorch: An imperative style, high-performance deep learning library.
\newblock \emph{arXiv preprint arXiv:1912.01703}, 2019.

\bibitem[Pinto et~al.(2017)Pinto, Andrychowicz, Welinder, Zaremba, and
  Abbeel]{pinto2017asymmetric}
L.~Pinto, M.~Andrychowicz, P.~Welinder, W.~Zaremba, and P.~Abbeel.
\newblock Asymmetric actor critic for image-based robot learning, 2017.

\bibitem[Radford et~al.(2019)Radford, Wu, Child, Luan, Amodei, and
  Sutskever]{gpt2}
A.~Radford, J.~Wu, R.~Child, D.~Luan, D.~Amodei, and I.~Sutskever.
\newblock Language models are unsupervised multitask learners.
\newblock \emph{OpenAI blog}, 1\penalty0 (8):\penalty0 9, 2019.

\bibitem[Schulman et~al.(2017)Schulman, Wolski, Dhariwal, Radford, and
  Klimov]{ppo}
J.~Schulman, F.~Wolski, P.~Dhariwal, A.~Radford, and O.~Klimov.
\newblock Proximal policy optimization algorithms, 2017.

\bibitem[Silver et~al.(2016)Silver, Huang, Maddison, Guez, Sifre, Van
  Den~Driessche, Schrittwieser, Antonoglou, Panneershelvam, Lanctot,
  et~al.]{alphago}
D.~Silver, A.~Huang, C.~J. Maddison, A.~Guez, L.~Sifre, G.~Van Den~Driessche,
  J.~Schrittwieser, I.~Antonoglou, V.~Panneershelvam, M.~Lanctot, et~al.
\newblock Mastering the game of go with deep neural networks and tree search.
\newblock \emph{nature}, 529\penalty0 (7587):\penalty0 484--489, 2016.

\bibitem[Smith(2017)]{cyclical}
L.~N. Smith.
\newblock Cyclical learning rates for training neural networks, 2017.

\bibitem[Tesauro(1995)]{selfplay}
G.~Tesauro.
\newblock Temporal difference learning and td-gammon.
\newblock \emph{Communications of the ACM}, 38\penalty0 (3):\penalty0 58--68,
  1995.

\bibitem[Vaswani et~al.(2017)Vaswani, Shazeer, Parmar, Uszkoreit, Jones, Gomez,
  Kaiser, and Polosukhin]{transformer}
A.~Vaswani, N.~Shazeer, N.~Parmar, J.~Uszkoreit, L.~Jones, A.~N. Gomez,
  L.~Kaiser, and I.~Polosukhin.
\newblock Attention is all you need, 2017.

\bibitem[Wichrowska et~al.(2017)Wichrowska, Maheswaranathan, Hoffman,
  Colmenarejo, Denil, Freitas, and Sohl-Dickstein]{scale_generalize}
O.~Wichrowska, N.~Maheswaranathan, M.~W. Hoffman, S.~G. Colmenarejo, M.~Denil,
  N.~Freitas, and J.~Sohl-Dickstein.
\newblock Learned optimizers that scale and generalize.
\newblock In \emph{International Conference on Machine Learning}, pages
  3751--3760. PMLR, 2017.

\bibitem[Wolf et~al.(2019)Wolf, Debut, Sanh, Chaumond, Delangue, Moi, Cistac,
  Rault, Louf, Funtowicz, et~al.]{huggingface}
T.~Wolf, L.~Debut, V.~Sanh, J.~Chaumond, C.~Delangue, A.~Moi, P.~Cistac,
  T.~Rault, R.~Louf, M.~Funtowicz, et~al.
\newblock Huggingface's transformers: State-of-the-art natural language
  processing.
\newblock \emph{arXiv preprint arXiv:1910.03771}, 2019.

\bibitem[Wu et~al.(2018)Wu, Ren, Liao, and Grosse]{short_horizon_bias}
Y.~Wu, M.~Ren, R.~Liao, and R.~Grosse.
\newblock Understanding short-horizon bias in stochastic meta-optimization,
  2018.

\bibitem[Xiao et~al.(2017)Xiao, Rasul, and Vollgraf]{fashionmnist}
H.~Xiao, K.~Rasul, and R.~Vollgraf.
\newblock Fashion-mnist: a novel image dataset for benchmarking machine
  learning algorithms, 2017.

\bibitem[Xu et~al.(2017)Xu, Qin, Wang, and Liu]{xu2017reinforcement}
C.~Xu, T.~Qin, G.~Wang, and T.-Y. Liu.
\newblock Reinforcement learning for learning rate control, 2017.

\bibitem[Xu et~al.(2019)Xu, Dai, Kemp, and Metz]{xu2019learning}
Z.~Xu, A.~M. Dai, J.~Kemp, and L.~Metz.
\newblock Learning an adaptive learning rate schedule.
\newblock \emph{arXiv preprint arXiv:1909.09712}, 2019.

\bibitem[You et~al.(2017)You, Gitman, and Ginsburg]{lars}
Y.~You, I.~Gitman, and B.~Ginsburg.
\newblock Large batch training of convolutional networks, 2017.

\bibitem[You et~al.(2020)You, Li, Reddi, Hseu, Kumar, Bhojanapalli, Song,
  Demmel, Keutzer, and Hsieh]{lamb}
Y.~You, J.~Li, S.~Reddi, J.~Hseu, S.~Kumar, S.~Bhojanapalli, X.~Song,
  J.~Demmel, K.~Keutzer, and C.-J. Hsieh.
\newblock Large batch optimization for deep learning: Training bert in 76
  minutes, 2020.

\bibitem[Zhang et~al.(2019)Zhang, Li, Nado, Martens, Sachdeva, Dahl, Shallue,
  and Grosse]{nqm}
G.~Zhang, L.~Li, Z.~Nado, J.~Martens, S.~Sachdeva, G.~E. Dahl, C.~J. Shallue,
  and R.~Grosse.
\newblock Which algorithmic choices matter at which batch sizes? insights from
  a noisy quadratic model, 2019.

\end{thebibliography}

\newpage

\begin{appendices}

\section{Scaling Laws for Large Language Models}

\begin{figure}[h]
    \centering
    \includegraphics[scale=0.38]{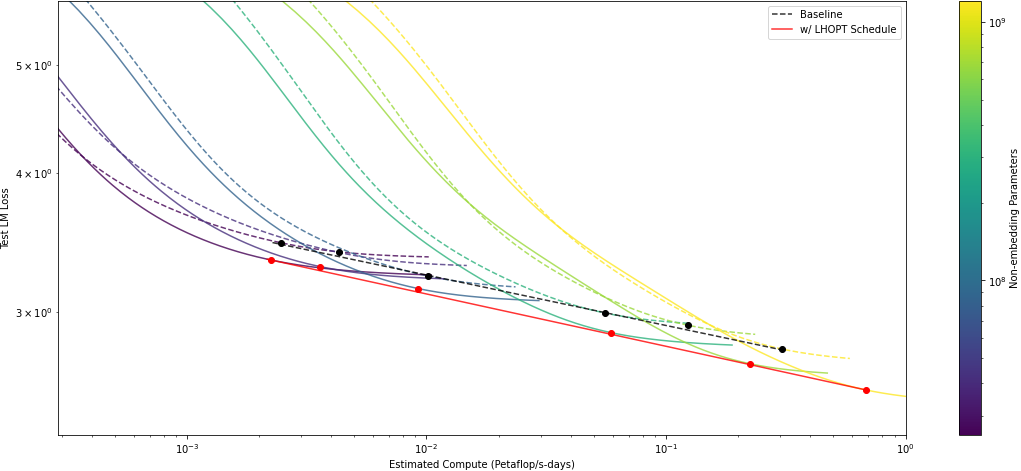}
    \caption{Same data as Figure \ref{fig:lm_speedup} but also including parameter count colorbar.}
    \label{fig:lm_speedup_old}
\end{figure}

\section{Fixed Schedule for Language Modeling}
\label{appendix:fixed_schedule}

\begin{figure}[h]
    \centering
    \includegraphics[scale=0.6]{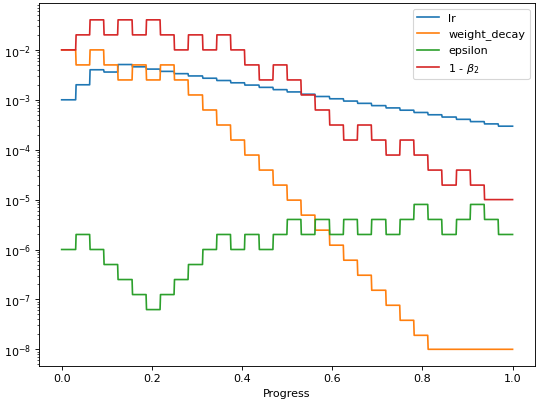}
    \caption{Hyperparameter settings for the fixed language modeling schedule.}
    \label{fig:fixed_schedule}
\end{figure}

Figure \ref{fig:fixed_schedule} shows the values of the 4 hyperparameters over time used for all model sizes. $\beta_1$ was also available as a hyperparameter for the inner optimizer, but our experiments showed $\beta_1$ to be difficult to set robustly from small models - in particular, we had found that the ideal $\beta_1$ was a function of transformer context length and thus just kept it set at the default value (0.9).

\section{Why Use the LAMB Trust Ratio}
\label{appendix:why_lamb}

Basing our inner optimizer on LAMB may be seen as an unusual choice, especially because the LAMB optimizer has mostly been advertised as for training with very large batches~\citep{lamb} while instead our inner training is on incredibly small models and batch sizes.

The reason we chose LAMB is, like many other of our design decisions, for generalization. For optimizers in the family as Adam~\citep{adam}, the numerator and denominator cancel out gradient scale and we are left with updates of approximately the same size as the learning rate. The issue is that past research has found that parameter scale is quite important to optimization~\citep{glorot} and that this scale is a function of network width. Reasonably, past work has also chosen to set learning rate as a function of model size~\citep{scaling_laws}.

These issues combined make it difficult for learned optimizers to robustly set learning rates for Adam-like optimizers because it would involve them having to learn to extrapolate beyond the training set. LAMB-like optimizers do not have this issue.

While the above is the main reason for using LAMB, there are several other benefits as well: (1) Even if one set the learning rate optimally based on the width of the network, there are some parameters, such as biases and normalization scales, which don't have this property and setting learning rate in this way effectively slows down the learning of these parameters as model scale increases. (2) Having hyperparameters behave independently is a useful property for being able to control optimizer behavior, and while past research has noted the importance of $\epsilon$ for Adam~\citep{empirical_comparisons}, extremely large learning rates need to be used with it. For Adam, increasing both learning rate and $\epsilon$ together allow the latter to dominate the second moment term in the denominator and simply behave like SGD-M. For LAMB, $\epsilon$ can be set independently, and as it approaches infinity, the optimizer updates simply approach LARS~\citep{lars} - no learning rate tuning needed.

\section{Do LHOPTs Overfit The Validation Set?}
\label{appendix:overfit}

\begin{figure}[h]
    \centering
    \includegraphics[scale=0.6]{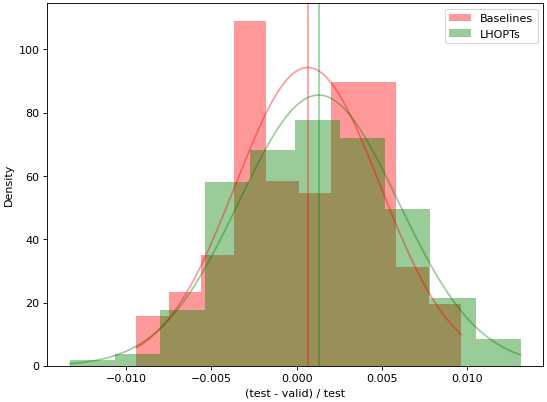}
    \caption{Showing the distribution of relative difference between test and validation loss on ResNet18. We find that LHOPTs do not significantly overfit to the validation set.}
    \label{fig:overfitting_loss}
\end{figure}

Since they take in validation loss, one potential concern is overfitting that validation set and requiring an additional set for evaluating models trained with LHOPTs. In order to investigate this, we look at the distributions of the relative difference between test and validation scores for ResNet18. This dataset-model combination was chosen because we had the most data for baseline and LHOPT runs. We find that despite theoretically being able to overfit the validation set, these models generally do not for loss ($p=0.14$) (see Figure \ref{fig:overfitting_loss}), top-1 accuracy ($p=0.68$), or top-5 accuracy ($p=0.36$) and the relative differences generally appear to be from the same distribution. For reference, comparing AdamW to SGD shows that SGD overfits signficiantly more at all 3 metrics: loss ($p=1.49 \times 10^{-22}$), top-1 accuracy ($p=7.32 \times 10^{-7}$), and top-5 accuracy ($p=0.04$).

One limitation of this analysis is that we look at the entire curve, and as we have noted, our models tend to solely optimize for final performance, and thus the overfitting might be crowded out by not doing so before the final point(s). This may still be possible, but the $p$-values are even higher when looking at the final data points ($p>0.8$), but this may be due to having much fewer data points for comparing distributions.

Note that the $p$-values are similar if we look at absolute difference between test and valid metrics.

\section{Integral CDF Features}
\label{appendix:integral_cdf}

Given a mean $\mu$ and standard deviation $\sigma$, it's a single line of code to convert a value into a z-score and apply a cdf to that z-score (As mentioned in Section \ref{feature_space}). One issue is in calculating robust $\mu$ and $\sigma$. For example, if one used a global $\mu$ and $\sigma$, these statistics would change faster at the beginning of inner training, and slower at the end. Another possibility is using an exponential moving $\mu$ and $\sigma$, which solves the previous problem but introduces a new problem: that how quickly the statistics react is a function of how frequently the statistics are updated. For example, a point half-way through training would have much more weight on the final $\mu$ and $\sigma$ if we update the statistics twice as frequently (i.e., we double the number of outer steps). Our final solution to this is what we call Integral CDF Features: we replace the discrete sampling of points at each outer step by linearly interpolating between them, assign the weight of each point with the continuous version of an exponential moving average. Specifically, given observations $y_1, y_2, ..., y_N$ at progress values of $t_1, t_2, ..., t_N$, we would calculate weighted average $\mu$ as:

$$\mu = \frac{\sum_{n=1}^{N-1} \int_{t=t_n}^{t_{n+1}} e^{t * log(b)}[ y_n \frac{t - t_n}{t_{n+1} - t_n} + y_{n+1} \frac{t_{n+1} - t}{t_{n+1} - t_n} ] dt}{\int_{t=t_1}^{t_{N}} e^{t * log(b)} dt}$$

And using the same weights for $\sigma$. $b$ here is specified as a hyperparameter corresponding to how much more weight a feature at the end of inner training ($t=1$) has than one at the beginning of training ($t=0$). Rather than tune $b$, we instead make a copy for each feature for several different values of $b$. For this paper, we use $b \in [1.25, 2.5, 5, 10, 20]$.

\section{ImageNet w/ SGD}
\label{appendix:old_imagenet}

When we first started this project, our earlier models were inner optimizer-agnostic and as part of the task, we also chose a random inner optimizer between SGD/Adam/Adamax. These models actually had very strong performance, outperforming our newer ones on ImageNet (Table \ref{old_imagenet_table}) and beating learning rate and schedule tuned SGD on the task.

The reason we moved away from these models and instead chose to specialize on an inner optimizer was because this allow us to control more hyperparameters, and the problems that require the largest use of compute generally use adaptive optimizers.

\begin{figure}[h]
    \begin{subfigure}{.5\textwidth}
        \includegraphics[scale=0.39]{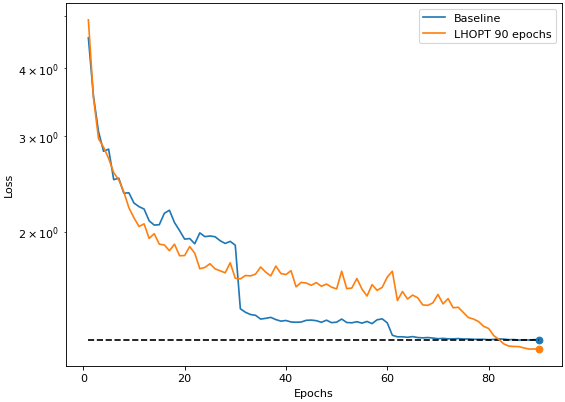}
        \caption{}
        \label{fig:old_resnet18}
    \end{subfigure}%
    \begin{subfigure}{.5\textwidth}
        \includegraphics[scale=0.39]{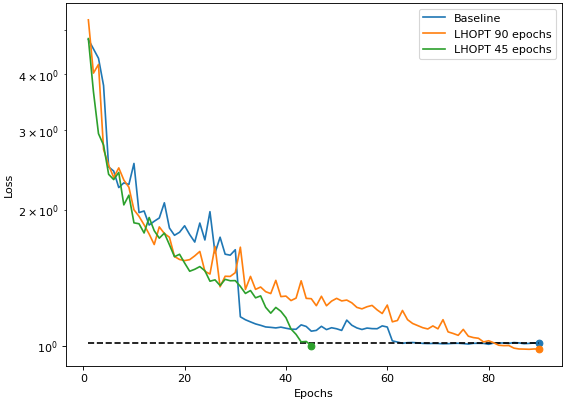}
        \caption{}
        \label{fig:old_resnet50}
    \end{subfigure}
    \caption{Test learning curve of a tuned SGD baseline and older versions of our LHOPT that were trained to be robust to inner optimizer on (\ref{fig:old_resnet18}) Resnet18 and (\ref{fig:old_resnet50}) ResNet50.}
    \label{fig:old_imagenet}
\end{figure}

\begin{table}
  \caption{Imagenet Performance using SGD.}
  \label{old_imagenet_table}
  \centering
  \begin{tabular}{lllll}
    \toprule
    Model & Epochs & Acc@1 & Acc@5 & Loss \\
    \midrule
    Resnet18 & 90 & 69.10 & 88.62 & 1.26 \\
    Resnet18 + LHOPT & 90 & $\mathbf{69.97}$ & $\mathbf{89.31}$ & $\mathbf{1.21}$ \\
    \midrule 
    Resnet50 & 90 & 74.66 & 92.09 & 1.01 \\
    Resnet50 + LHOPT & 45 & 74.79 & 92.18 & 1.00 \\
    Resnet50 + LHOPT & 90 & $\mathbf{75.50}$ & $\mathbf{92.49}$ & $\mathbf{0.98}$ \\
    \bottomrule
  \end{tabular}
\end{table}

\section{How Important are Larger Models?}

One question we had was if scaling up our controller would be beneficial, since our default one was so small. We show results on Table \ref{traindist_table} for a large version of the controller that uses 2 hidden LSTM layers and increasing with to 512. We can see from that table that the larger model performs better on the distribution of tasks than the default, but from the limited testing we performed, the larger model seemed to perform quite similarly on real-world tasks and thus don't include the results. One possible reason for this could be that the larger models could be doing better at tasks with unusual settings in the task distribution, but doing better at those may not apply to real world tasks.

\section{Inner Task Distribution}
\label{appendix:training_distribution}

Table \ref{training_distribution} shows the breakdown of our training distribution. None of the task ratios were tuned - the unusual values came from incrementally adding more datasets to the mix over time. Performance rarely improved when adding new datasets, but we chose to do so anyway because our prior is that they would be helpful. Section \ref{task_diversity} was our attempt at answering whether or not it was important to do so retrospectively, and it seems like the additional inner tasks were not very helpful.

The full domain specific language for our training distribution is available at \url{https://github.com/openai/LHOPT}, but we provide some additional high-level details: RNN in the table refers to randomly using a simple RNN, LSTM, or GRU~\citep{gru}. For CNNs and MLPs, we randomize over 5 different activation functions (ReLU, leaky ReLU, very leaky ReLU, ELU, PReLU) and normalizations (none, BN, LN). For MLPs, we additionally randomize over loss functions (CCE, MAE, MSE, Huber). Algo + RNN is on XOR and binary addition tasks. Text + RNN is classification on the SST dataset. Text + Transformer is language modeling over 31 different simple datasets.

\begin{table}
  \caption{Inner Task Distribution.}
  \label{training_distribution}
  \centering

    \begin{tabular}{lr}
    \toprule
    {} &  Ratio \\
    \midrule
    Algo + RNN         &   0.27 \\
    MNISTs + CNN       &  0.068 \\
    MNISTs + MLP       &    0.2 \\
    NQM                &  0.014 \\
    Text + RNN         &   0.14 \\
    Text + Transformer &   0.31 \\
    \bottomrule
    \end{tabular}
\end{table}

\section{Initial Inner Hyperparameters, Noise, and Actions}
\label{appendix:inner_optimizer}

\subsection{Initial Inner Hyperparameters}

Table \ref{initial_inner_hyperparameters} shows the initial hyperparameter. Note that we normally represent moving average rates as $1 - \beta$ to make multiplicative scaling actions easier to apply. We also include several novel hyperparameters:

\begin{enumerate}
\item Grad clip fraction: instead of gradient clipping to an absolute value (which cannot be robust to task), we instead clip gradients as a fraction of an exponentially weighted moving maximum of gradient norms.
\item $\beta_{gradclip}$: this is the rate at which the maximum decays: $G_t = max(G_{t-1} * \beta_{gradclip}, ||grad_t||)$
\item $\beta_{LAMB}$: Instead of normalizing by the update norm in LAMB, we instead normalize by an exponential moving average of the update norm - which allows us to recover relative update norm information.
\end{enumerate}

\begin{table}[h]
  \caption{Initial Inner Hyperparameters.}
  \label{initial_inner_hyperparameters}
  \centering
    \begin{tabular}{lr}
    \toprule
    Hyperparameter &  Value \\
    \midrule
    Learning rate & $1e^{-3}$ \\
    $1 - \beta_1$ & 0.1 \\
    $1 - \beta_2$ & $1e^{-2}$ \\
    $\epsilon$ & $1e^{-6}$ \\
    Weight decay & $1e^{-2}$ \\
    Grad clip fraction & 0.8 \\ 
    $1 - \beta_{gradclip}$ & $1e^{-2}$ \\
    Denominator & Adamax \\
    Use LAMB & True \\
    LAMB minimum trust~\citep{sharnn} & $1e^{-3}$ \\
    $1 - \beta_{LAMB}$ & $0.05$ \\
    \bottomrule
    \end{tabular}
\end{table}

\subsection{Actions and Noise}

In addition to the actions in Table \ref{actions_noise}, we also had restart actions corresponding to 3 checkpoints with 3 actions per checkpoint: save current state to the checkpoint, load state at the checkpoint, and swap states between current and checkpoint. This results in a total of 10 restart actions (3 * 3 + 1 for doing nothing).

In the table, noise is represented by an upper and lower bound that is applied by the corresponding action type. For example, initial learning rate is scaled by some value sampled in [$1e^{-2}$,  $1e^{2}$] when the task is sampled.

\begin{table}[h]
  \caption{Actions and Noise.}
  \label{actions_noise}
  \centering
    \begin{tabular}{lrrr}
    \toprule
    Hyperparameter &  Action Type & Action Values & Noise \\
    \midrule
    Learning rate & Scale & [0.5, 0.707, 0.9, 1, 1.1, 1.414, 2.0] & [$1e^{-2}$,  $1e^{2}$] \\
    Weight decay & Scale & [0.5, 2.0] & [0.1, 10] \\
    $\epsilon$ & Scale & [0.5, 2.0] & [0.1, 10] \\
    $1 - \beta_1$ & Scale & [0.5, 2.0] & [0.1, 10] \\
    $1 - \beta_2$ & Scale & [0.5, 2.0] & [0.1, 10] \\
    Grad clip fraction & Logit Shift & [-1, -0.3, 0.3, 1] & [-1, 1] \\ 
    $1 - \beta_{gradclip}$ & Scale & [0.5, 1.0, 2.0] & [0.5, 2] \\
    $1 - \beta_{LAMB}$ & Scale & [$\frac{1}{1.5}$, 1.0, 1.5] & [0.5, 2] \\
    \bottomrule
    \end{tabular}
\end{table}

\section{$\pi$ and $V$ Features}
\label{appendix:features}

\subsection{$\pi$ Features}

For some features, we will mention transforms applied to them. is\_nan refers to a feature on whether or not that value is NaN. NaN's are all converted to a default value to avoid passing them in to the controller. CDF features refers to the 5 features in Appendix \ref{appendix:integral_cdf}.

Features updated at each outer step:
\begin{enumerate}
\item Progress $\in [0,1]$
\item Previous action representation
\item Log loss ratio (train and valid): is\_nan, tanh, CDF features 
\item Log train loss: is\_nan, is\_inf, CDF features 
\item Log valid loss: is\_nan, is\_inf, CDF features
\item Percentile of loss at each checkpoint
\item Progress at each checkpoint
\item For all hyperparameter actions: current value / initial value
\item Log ratio of previous and current param norm: tanh, CDF features
\item Log ratio of update and previous param norm: tanh, CDF features
\end{enumerate}

Features updated every 4 inner steps (for all of these features, we include the raw value as well as the CDF features):
\begin{enumerate}
\item Fraction of gradients clipped
\item Fraction of $\sqrt{\hat{v}} \geq \epsilon$
\item Average per-parameter update magnitude (pre-LR)
\item Log scale from~\citep{2regimes}
\item Log ratio of update norm and param norm
\item Log of noise scale~\citep{mccandlish2018empirical}
\item Log LAMB trust ratio
\item Cosine similarity of gradient and momentum
\item Logit of CDF cosine similarity of gradient and momentum
\item Cosine similarity of gradient and update
\item Logit of CDF cosine similarity of gradient and update
\item Cosine similarity of gradient and parameter
\item CDF cosine similarity of gradient and parameter
\end{enumerate}

One issue with cosine similarity is that it is not robust to vector size: the expected cosine similarity between random vectors is a function of their size. In order to make this more robust, we additionally introduce a "CDF cosine similarity", where we apply a gaussian CDF to a cosine similarity times the square root of the dimensionality of the vector. We additionally apply the logit function on top of this when we expect the value to be very close to 1.

\subsection{$V$ Features}

In addition to all the features for $\pi$, we pass in the following additional features to our value function:
\begin{enumerate}
\item A vector representation of the task: all high-level task choices are converted into a set of values. For example, if we allow between 1 and 7 layers for a MLP and the task chooses 4, this would be encoded as a one-hot vector.
\item A vector representation of the random initial hyperparameters
\item Reward
\item Log train and eval loss
\item Statistics about the reward baseline optimizer's losses (From Section \ref{reward_function})
\end{enumerate}
\end{appendices}

\end{document}